\documentclass[]{spie}

\usepackage{amsmath,amsfonts,amssymb}
\usepackage{graphicx}
\usepackage{booktabs}
\usepackage[colorlinks=true, allcolors=blue]{hyperref}

\title{A Probabilistic Segment Anything Model for Ambiguity-Aware Medical Image Segmentation}

\author{Tyler Ward}
\author{Abdullah Imran}
\affil{Department of Computer Science\\University of Kentucky, Lexington, KY 40506, USA}

\begin{document} 
\maketitle

\begin{abstract}
Recent advances in promptable segmentation, such as the Segment Anything Model (SAM), have enabled flexible, high-quality mask generation across a wide range of visual domains. However, SAM and similar models remain fundamentally deterministic, producing a single segmentation per object per prompt, and fail to capture the inherent ambiguity present in many real-world tasks. This limitation is particularly troublesome in medical imaging, where multiple plausible segmentations may exist due to annotation uncertainty or inter-expert variability. In this paper, we introduce \textit{Probabilistic SAM}, a probabilistic extension of SAM that models a distribution over segmentations conditioned on both the input image and prompt. By incorporating a latent variable space and training with a variational objective, our model learns to generate diverse and plausible segmentation masks reflecting the variability in human annotations. The architecture integrates a prior and posterior network into the SAM framework, allowing latent codes to modulate the prompt embeddings during inference. The latent space allows for efficient sampling during inference, enabling uncertainty-aware outputs with minimal overhead. We evaluate Probabilistic SAM on the public LIDC-IDRI lung nodule dataset and demonstrate its ability to produce diverse outputs that align with expert disagreement, outperforming existing probabilistic baselines on uncertainty-aware metrics. Our code is available at: \url{https://github.com/tbwa233/Probabilistic-SAM/}.
\end{abstract}

\section{Introduction}
\label{sec:intro}

Semantic segmentation is a fundamental task in computer vision, with wide-ranging applications from autonomous driving to medical diagnostics. In many real-world scenarios, particularly in the medical domain, image annotations are inherently ambiguous in that different experts may provide different segmentations of the same image \cite{schmarje2023annotating}, reflecting true uncertainty rather than labeling error. Traditional segmentation models, however, are designed to produce a single deterministic output. This approach ignores the diversity of valid interpretations and fails to capture the uncertainty present in the data \cite{czolbe2021segmentation}. As a result, such models may produce overconfident predictions and struggle in domains where multiple plausible outcomes exist.

Promptable segmentation foundation models like the Segment Anything Model (SAM)~\cite{kirillov2023segment} have introduced a new paradigm by allowing user guidance in the form of points, boxes, or masks. Prompting, while powerful, does carry some inherent limitations, chief among them being that it is difficult to design scalable systems based on manual prompting~\cite{ward2025annotation}. While recent work has proposed methods to automate the prompting process of SAM in a more annotation-efficient manner \cite{ward2025detection, li2025autoprosam, ward2025autoadaptive}, SAM remains deterministic and does not account for variability in user prompts or inherent ambiguity in the underlying data~\cite{ma2024segment, li2024flaws}. In domains like medical imaging, where lesion boundaries are subjective and clinician prompts may differ, the inability to model uncertainty can significantly limit the model's reliability and interpretability~\cite{lambert2024trustworthy}.

To address these limitations, we propose \textit{Probabilistic SAM}, a novel extension of SAM that introduces probabilistic reasoning into the segmentation process. Specifically, our contributions are as follows:

\begin{itemize}
    \item We propose \textit{Probabilistic SAM}, a novel adaptation of SAM that introduces a learned latent space for modeling segmentation uncertainty.
    \item The model uses a posterior and prior network to sample latent representations, enabling it to generate diverse masks through variational training.
    \item We demonstrate the model’s ability to capture annotator variation and inherent ambiguity in segmenting lung nodules from chest computed tomography (CT) images.
\end{itemize}

\section{Methods}
\label{sec:methods}

\begin{figure}
\centering
\includegraphics[width=\linewidth]{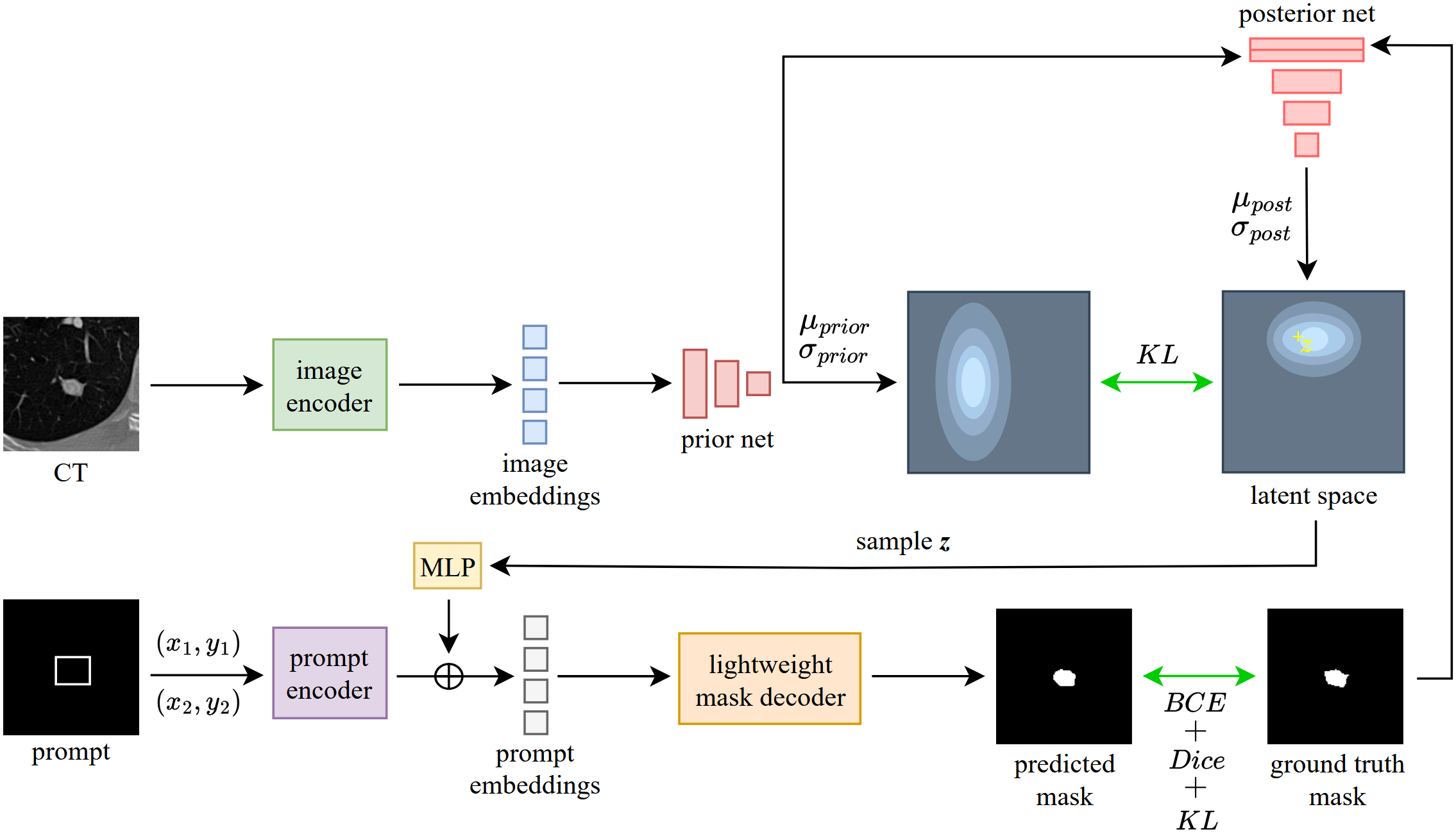}
\caption{The training procedure of Probabilistic SAM. Given a CT slice and a bounding box prompt $(x_1, y_1), (x_2, y_2)$, visual and spatial information is encoded via SAM's image and prompt encoders. During training, a posterior network uses image embeddings and the ground truth mask to estimate $\mathcal{N}(\mu_{\text{post}}, \sigma_{\text{post}})$ while a prior network predicts $\mathcal{N}(\mu_{\text{prior}}, \sigma_{\text{prior}})$. A latent vector $z\sim\mathcal{N}(\mu_{\text{post}}, \sigma_{\text{post}}))$ sampled from the posterior network is projected and added to the prompt embeddings before decoding. The model is optimized using a combination of binary cross-entropy (BCE), Dice loss, and Kullback-Leibler (KL) divergence between the posterior and prior distributions.}
\label{fig:training}
\end{figure}

Our proposed Probabilistic SAM extends SAM by utilizing conditional variational autoencoders (CVAEs)~\cite{sohn2015learning}, enabling it to model a distribution over plausible segmentations conditioned on both the input image and a given prompt. By combining a latent variable model with SAM’s powerful prompt-based segmentation pipeline, Probabilistic SAM can capture both inherent image uncertainty and ambiguity in prompt interpretation. Our methodology consists of two main stages: training and sampling.

Fig.~\ref{fig:training} illustrates the training procedure of our proposed Probabilistic SAM. During training, Probabilistic SAM learns to model the distribution over plausible segmentations by leveraging variational inference~\cite{kingma2013auto}. The model optimizes the evidence lower bound objective (ELBO) by training a prior network $p(z | x)$ and a posterior network $q(z | x, y)$, where $x$ is the input image and $y$ is its corresponding ground truth mask. Both networks encode their respective inputs into a latent distribution over $z$, a vector capturing uncertainty and ambiguity in segmentation.

\begin{figure}
\centering
\includegraphics[width=\linewidth]{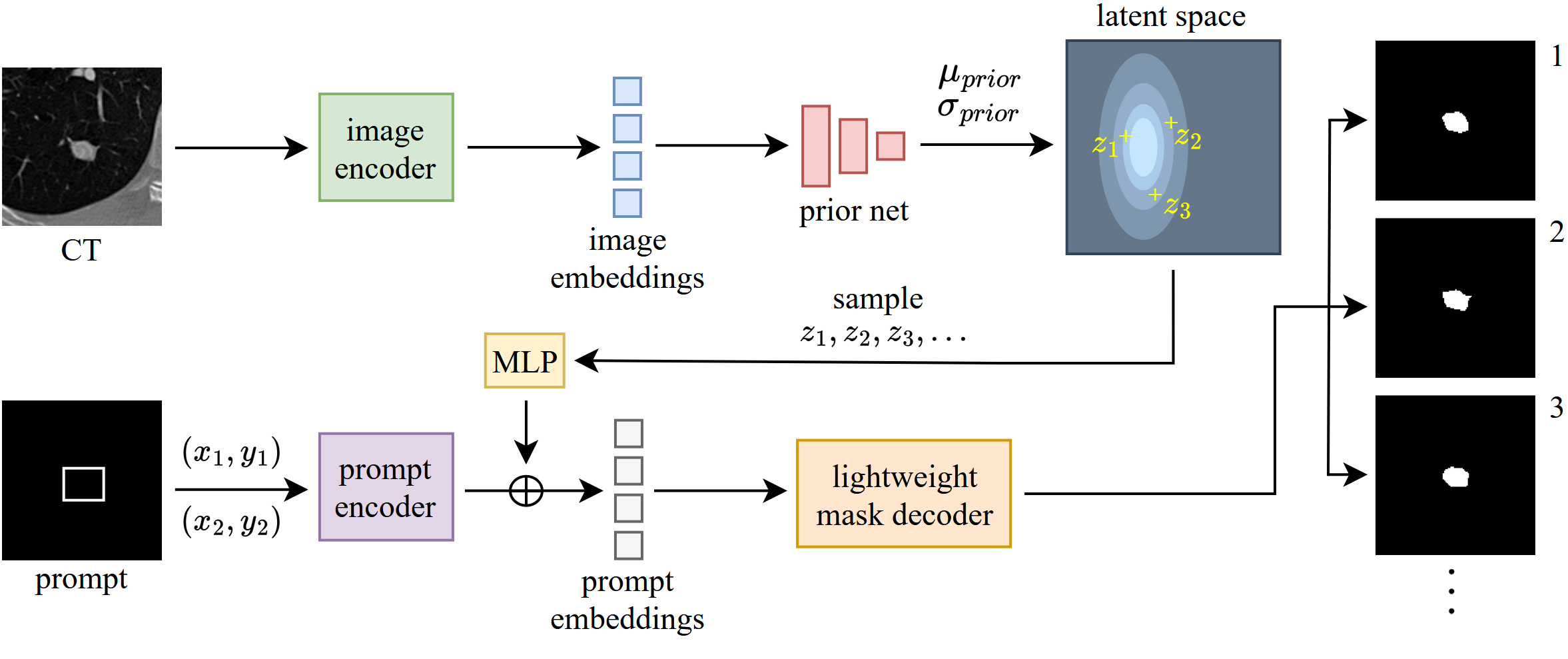}
\caption{The sampling process of Probabilistic SAM. A prior network maps image embeddings to a Gaussian latent space, from which latent vectors $z_1, z_2, z_3, \dots$ are sampled. After projection through a multilayer perceptron (MLP), these vectors are added to the sparse prompt embeddings. The modified prompts and image embeddings are passed to SAM's lightweight mask decoder to generate diverse segmentation predictions.}
\label{fig:sampling}
\end{figure}

The posterior network receives both the image and ground truth mask as input and produces a distribution $q(z | x, y)$ from which a latent vector $z$ is sampled. This sample is projected via an MLP and added to the prompt embeddings before being passed to the SAM mask decoder, along with image embeddings and positional embeddings. Then the predicted mask is obtained for the prompted object-of-interest. The predicted mask is compared to the ground truth using a reconstruction loss consisting of a combination of BCE and Dice loss:

\begin{equation}
\mathcal{L}_{\text{recon}} = \mathcal{L}_{\text{BCE}}(y, \hat{y}) + \mathcal{L}_{\text{Dice}}(y, \hat{y}),
\end{equation}

\noindent where $\hat{y}$ is a predicted mask. Individually, the BCE and Dice losses are expressed as:

\begin{equation}
\mathcal{L}_{\text{BCE}}(y, \hat{y}) = -\frac{1}{N}\sum_{i=1}^{N}[y_i \log \hat{y}_i+(1-y_i) \log (1-\hat{y}_i],
\end{equation}

\begin{equation}
\mathcal{L}_{\text{Dice}}(y, \hat{y})=1-\frac{2\sum_iy_i\hat{y}_i+\epsilon}{\sum_iy_i+\sum_i\hat{y}_i+\epsilon}.
\end{equation}

Meanwhile, the KL divergence, $D_{\text{KL}}[q(z | x, y) \| p(z | x)]$, encourages alignment between the learned posterior and prior distributions. The total loss is the sum of the reconstruction loss and KL divergence, weighted by a hyperparameter $\beta$. $\beta$ is set to 10 in our implementation.

\begin{equation}
\mathcal{L}_{\text{total}} = \mathcal{L}_{\text{recon}} + \beta \cdot D_{\text{KL}}[q(z | x, y) \| p(z | x)].
\end{equation}

\noindent This procedure allows the model to learn a structured latent space over segmentation variability, conditioned on both image content and prompting information.

At inference time (see Fig.~\ref{fig:sampling}), Probabilistic SAM generates diverse plausible segmentation masks by sampling from a learned prior distribution over a latent variable $z$. Given an input image, the image encoder from the original SAM architecture first produces a high-quality image embedding. Separately, the prompt encoder processes user-provided prompts to produce corresponding sparse and dense prompt embeddings. 

A latent vector $z\sim p(z | x)$ is sampled from the prior network, which conditions only on the image. This sampled latent variable is projected via a small MLP and added to the sparse prompt embedding. These modified prompt embeddings, along with the original dense prompt embeddings and image embeddings, are then passed to SAM's frozen mask decoder to generate the final segmentation mask. By repeating this sampling process multiple times with different latent samples, the model can produce a diverse set of plausible masks, reflecting the ambiguity or subjectivity inherent in the segmentation task.

\section{Experiments and Results}
\label{sec:results}

\subsection{Data}
\label{subsec:data}
We validated our proposed Probabilistic SAM on the Lung Image Database Consortium and Image Database Resource Initiative (LIDC-IDRI) dataset\cite{armato2011lung}. This dataset contains lesion annotations collected from four expert radiologists across 1,018 lung CT scans from 1,010 patients. Following Kohl et al.~\cite{kohl2018probabilistic}, we resample the CTs to 0.5 mm $\times$ 0.5 mm in-plane resolution and cropped 2D slices centered at the position of the lesion. We then split the dataset into train/val/test sets, consisting of 722 scans (8,882 slices) for the training set, and 144 scans (1,996 slices) each for the validation and testing sets.

\subsection{Implementation Details}
\label{subsec:implementation}
All experiments were performed on an \emph{Intel (R) Xeon (R) w7-2475X, 2600MHz} machine with a dual \emph{NVIDIA A4000X2} GPU (32GB). To gauge the performance of Probabilistic SAM against existing probabilistic segmentation methods, we compare against Dropout U-Net and Probabilistic U-Net~\cite{kohl2018probabilistic}, and a Dropout SAM model constructed by applying dropout with probability $p=0.5$ in the transformer layers and output MLPs of SAM's mask decoder. For quantitative evaluation, we report the generalized energy distance (GED)~\cite{szekely2013energy}, Dice-Sørensen coefficient (DSC), and intersection-over-union (IoU) of each model.

\begin{figure}
    \centering
    {
    \vspace{1.2em}
    \begin{tabular}{c c c c c}
     & \multicolumn{4}{c}{Experts 1 $\rightarrow$ 4 Annotations}\\[0.2em]
     &
     \includegraphics[width=0.15\linewidth]{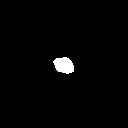}
     &
     \includegraphics[width=0.15\linewidth]{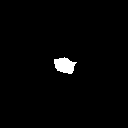}
     &
     \includegraphics[width=0.15\linewidth]{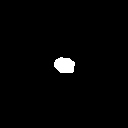}
     &
     \includegraphics[width=0.15\linewidth]{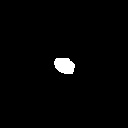}
     \\
     {CT} & \multicolumn{4}{c}{Probabilistic U-Net Outputs}\\[0.2em]
     \includegraphics[width=0.15\linewidth]{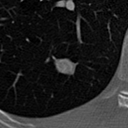}
     &
     \includegraphics[width=0.15\linewidth]{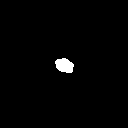}
     &
     \includegraphics[width=0.15\linewidth]{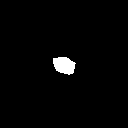}
     &
     \includegraphics[width=0.15\linewidth]{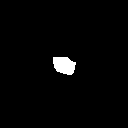}
     &
     \includegraphics[width=0.15\linewidth]{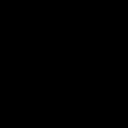}
     \\
     & \multicolumn{4}{c}{Probabilistic SAM Outputs}\\[0.2em]
     &
     \includegraphics[width=0.15\linewidth]{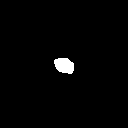}
     &
     \includegraphics[width=0.15\linewidth]{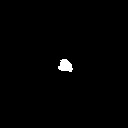}
     &
     \includegraphics[width=0.15\linewidth]{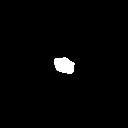}
     &
     \includegraphics[width=0.15\linewidth]{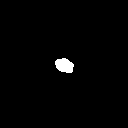}
     \vspace{1.0em}
    \end{tabular}
    }
    \caption{Qualitative results of Probabilistic SAM vs. Probabilistic U-Net and the ground truth annotations from four expert annotators (top row).}
    \label{fig:prob_masks}
\end{figure}

\subsection{Results}
Table~\ref{tab:quantitative} reports the quantitative performance of our Probabilistic SAM in segmenting lung nodules from chest CT images. Comparing against the baseline probabilistic and dropout methods, Probabilistic SAM is found to be more effective in accurately predicting lung nodule masks as well as in modeling the segmentation uncertainty. Our model achieves a GED of 0.2910 and an IoU of 0.7849 compared to Probabilistic U-Net's 0.3349 and 0.5557, an improvement of 4.39\% and 22.92\%, respectively. Probabilistic SAM also achieves a higher DSC (0.8255) compared to Dropout SAM (0.6799), an improvement of 14.56\%. A one-tailed paired t-test reveals that our Probabilistic SAM is significantly better than the compared methods including Probabilistic U-Net, in terms of all the evaluation metrics ($p$-values $<$0.05 ). This demonstrates Probabilistic SAM's enhanced ability to model uncertainty. Qualitative comparison further confirms the superiority of our Probabilistic SAM model. Visual comparison in Fig.~\ref{fig:prob_masks} demonstrates that Probabilistic SAM predicted segmentation outputs are better aligned with the ground truth annotations compared to its Probabilistic U-Net counterpart.

\label{subsec:results}

\begin{table}
    \centering
    \caption{Quantitative evaluation of our proposed Probabilistic SAM against existing work on probabilistic segmentation.
    }
    \vspace{1.0em}
    \label{tab:quantitative}
    \begin{tabular}{l c c c}
        \toprule
        Model & GED ($\downarrow$) & DSC ($\uparrow$) & IoU ($\uparrow)$\\
        \midrule
        Dropout U-Net & 0.5156 & 0.5591 & 0.3880\\
        Dropout SAM & 0.5025 & \underline{0.6799} & 0.5150\\
        Probabilistic U-Net & \underline{0.3349} & 0.5818 & \underline{0.5557}\\
        Probabilistic SAM & \textbf{0.2910} & \textbf{0.8255} & \textbf{0.7849}\\
        \bottomrule
    \end{tabular}
\end{table}

\section{Conclusions}
\label{sec:conclusions}
In this work, we have presented \textit{Probabilistic SAM}, a novel extension of SAM that enables the generation of diverse and uncertainty-aware segmentations from a single image and prompt. We demonstrated this capability on the LIDC-IDRI dataset, where the model successfully captures annotator variability and outperforms the baseline probabilistic and dropout methods. Our future work will focus on validating its functionality to model ambiguity at multiple scales and exploring its efficacy on additional segmentation tasks.

\bibliography{references}
\bibliographystyle{spiebib}

\end{document}